\documentclass{article}
\usepackage{spconf,amsmath,amssymb,graphicx}
\usepackage{cite}
\usepackage{enumitem}
\usepackage{listings}
\usepackage{xcolor} 
\usepackage{tabularx} 
\usepackage{booktabs}

\lstdefinestyle{python}{
    language=Python,
    basicstyle=\footnotesize\ttfamily,
    keywordstyle=\color{blue},
    stringstyle=\color{red},
    commentstyle=\color{green},
    breaklines=true,
    showstringspaces=false
}

\title{DebFlow: Automating Agent Creation via Agent Debate}
\name{Jinwei Su$^{1}$, Yinghui Xia$^{2}$, Yiqun Duan, Jun Du, Jianuo Huang, Tianyu Shi$^{3}$, Lewei He$^{1\dagger}$\thanks{$^\dagger$ Corresponding \ Author:\ helewei@m.scnu.edu.cn}}
\address{$^{1}$ School of Artificial Intelligence, South China Normal University \\
$^{2}$ AutoAgents.ai,  $^{3}$University of Toronto
}
\begin{document}
%
\maketitle
\begin{abstract}
Large Language Model (LLM)-based agentic systems have demonstrated strong capabilities across various tasks with a static workflow designed with expert knowledge. 
Recently, researchers have attempted to automate the generation and optimization of these workflows using code-based representations. 
However, most existing approaches rely on single-model reasoning combined with coarse-grained feedback—e.g., binary success/iterative failure long raw-logs, making workflow construction unstable and computationally expensive.
To address these limitations, we introduce \emph{DebFlow}, a framework that employs \emph{multi-agent debate} to achieve more reliable workflow optimization. In DebFlow, multiple agents propose, critique, and refine candidate workflows through iterative debate, guided by \emph{fine-grained feedback} formalized on a directed-graph representation. 
By revisiting and refining pilot trajectories under this debate-driven process—analogous to selective pressures in natural evolution—DebFlow converges toward more efficient workflows. Extensive experiments on six benchmarks show that DebFlow consistently surpasses prior automated frameworks in both accuracy and inference efficiency.

\end{abstract}
\section{Introduction}

Large Language Models (LLMs) have demonstrated strong capabilities across domains such as code generation\cite{Shinn2023ReflexionLA}, data analysis\cite{Hong2024DataIA}, and decision-making\cite{10378628}. To extend these abilities into complex multi-step tasks, prior work has relied on manually crafted agentic workflows that orchestrate LLM actions. However, such human-designed solutions require extensive domain expertise, limit adaptability to new problem settings, and hinder scalability. As history in machine learning suggests, hand-designed heuristics are ultimately replaced by automated, learnable frameworks.

Current research aims to develop automated frameworks for discovering efficient agentic workflows, thus minimizing human intervention. ADAS\cite{Hu2024AutomatedDO} defines the entire agentic system in code. However, the efficiency limitations of the linear heuristic search algorithm of ADAS hinder its ability to generate effective workflows within a constrained number of iterations. AFlow \cite{Zhang2024AFlowAA} models the workflow as a series of interconnected LLM-invoking nodes, where each node corresponds to an LLM action, and the edges capture the logical structure, dependencies, and execution flow between these actions. AFLOW employs the Monte Carlo Tree Search (MCTS) algorithm to automatically optimize LLM agent designs. In the search process, Monte Carlo Tree Search (MCTS) often performs numerous redundant optimizations, leading to significant computational overhead. This inefficiency increases the overall cost of the search, as it spends excessive resources on exploring suboptimal or irrelevant branches in the decision tree. Consequently, the algorithm's performance can be hindered by this unnecessary expenditure of computational effort, impacting its scalability and effectiveness in large or complex problem spaces. This underscores the need for more cost-effective and efficient methods to automate the generation of agentic workflows.

\begin{figure*}[htbp]
    \centering
    \includegraphics[width=1\textwidth]{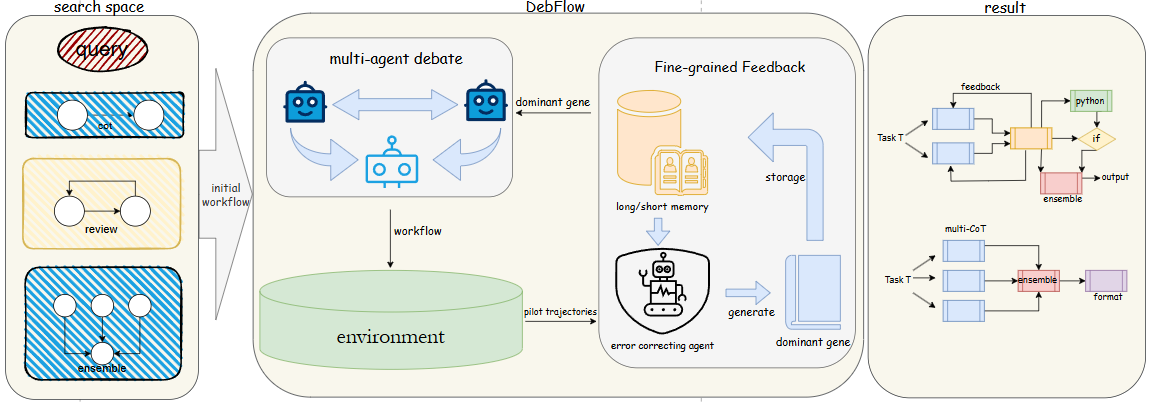} 
    \caption{The overall framework of \textbf{DebFlow.} The basic unit of framework invocation is the llm-invoking node, which can be combined to form different operators. The process starts with an initial workflow sampled from the search space. A multi-agent debate module generates candidate workflows, which interact with the environment and are evaluated. Fine-grained feedback analyzes failed workflows, stores key information, and guides subsequent debates by providing error-correcting signals. This iterative loop continues until the optimal workflow is obtained. Right: Case study and visualization. Tasks are from the MATH and HotpotQA benchmarks.}
    \label{total} 
\end{figure*}

Furthermore, previous works on ADAS\cite{Hu2024AutomatedDO} and AFlow \cite{Zhang2024AFlowAA} primarily comprised three core components: search space, search algorithm, and evaluation. In terms of search algorithms, these approaches predominantly relied on generating workflows through a single large language model (LLM), which significantly constrains the performance to the capabilities of the individual model.

In this work, we introduce \textbf{DebFlow}, a multi-agent framework that leverages role-playing debate to automate the generation and refinement of LLM-based workflows. Unlike prior debate-based methods~\cite{Du2023ImprovingFA,Wang2023UnleashingTE} that directly target query reasoning, DebFlow adapts debate to the domain of \textit{Automated Agentic Optimization}. Workflows are represented as directed graphs of reusable operators (e.g., Ensemble, Review-and-Revise), enabling compositionality. A novel fine-grained feedback mechanism analyzes workflow failures at the node and edge level, producing error-correcting signals that guide subsequent debates. This debate–feedback loop enables efficient exploration of workflow space while avoiding redundant search. Across six diverse benchmarks, DebFlow discovers agentic designs that consistently outperform prior automated frameworks and even surpass human-crafted baselines.
The key contributions of this work are as follows:
\begin{itemize}[label=\textbullet]
    \item We propose \textbf{DebFlow}, a framework that utilizes multi-agent debate to enhance the generation and optimization of workflows in large language model (LLM)-based agentic systems.
    \item \textbf{DebFlow} incorporates fine-grained feedback by revisiting pilot trajectories, enabling the system to extract informative signals that guide subsequent debates and workflow optimization.  
    \item Experiments on six representative tasks demonstrate that \textbf{DebFlow} consistently discovers workflow variants that achieve superior performance compared to current baselines.  
\end{itemize}

\section{Methodology}

In this section, we provide the technical details of \textbf{DebFlow}. The system overview is illustrated in Figure \ref{total}, where we utilize multi-agent debate and fine-grained feedback to facilitate automated workflow exploration.

\subsection{Problem Formulation}
We first define the basic unit of Debflow’s search space, namely the agentic operator as follows:

\textbf{Agentic Operator.} An agentic operator $\mathcal{O}$ is a composite LLM-agent invocation process that involves multiple LLM calls:
\begin{equation}\label{eq:operator}
\begin{gathered}
\mathcal{O} = \{\mathcal{M}_i, \mathcal{P}_i\, \mathcal{T}_i\},\quad\mathcal{M}_i \in \mathbb{M},\quad \mathcal{P}_i \in \mathbb{P},\quad \mathcal{T}_i \in \mathbb{T}
\end{gathered}
\end{equation}
where $\mathcal{M}$ and $\mathbb{M}$ correspond to LLM backbones and the set of available LLMs, respectively. Similarly, $\mathcal{P}$ and $\mathcal{T}$ represent prompts and tools.

\textbf{Workflow.} We define an \textbf{agentic workflow} as a directed acyclic graph (DAG):
\begin{equation}
\mathcal{W} = \{\mathcal{O}, \mathcal{E}\}, \quad \mathcal{O} \subset \mathbb{O}, \quad \mathcal{E} \subset \mathcal{O} \times \mathcal{O},
\end{equation}
where $\mathcal{O}$ denotes the set of selected operators and $\mathcal{E}$ encodes their directed connectivity. The DAG structure enforces hierarchical operator execution.

With this formulation, the workflow optimization problem can be expressed as:
\begin{equation}
\mathcal{W}^*=\arg\max_{W \in S} G(W, T),
\end{equation}
$\mathcal{W}^*$ is the optimal workflow configuration that maximizes the evaluation function $G$ for the given task $T$. $S$ is the search space.

\subsection{Multi-agent Debate}
Figure \ref{total} illustrates the Multi-Agent Debate. We use the form of multi-agent debate to make the model more directional when generating workflow, avoiding the "evolution" that failed in the past.

Given a task $T$, there are \( N \) \textbf{debaters}, denoted \( D = \{D_i\}_{i=1}^N \).
For each debater \( D = \{D_i\}_{i=1}^N \) in the debate round r, a history record $h_i$ will be maintained :
\begin{equation}
h_i= D_i(H_{r-1}, T, g, w)
\end{equation}
where $H$ represents the record of each history and $g$ is the dominant gene, generated by fine-grained feedback. $w$ is the workflow to be optimized. Update history: $H_r = H_{r-1} \cup \{h_1, h_2, \dots, h_N\}$.
To encourage thinking from more perspectives and improve the model’s reasoning ability, we use a role-playing approach. In the multi-agent debate, we assign roles of the affirmative and the opposing sides. Similar to a debate competition, one side presents its argument, while the other side refutes it and offers its own viewpoint.
Proponents propose a solution $S_p$:
\begin{equation}
    S_p = f_p(\{h_i \mid D_i \in \text{Proponents}\}, H_r, T, g, w)
\end{equation}
Opponents evaluate and propose a refined solution $S_o$:
\begin{equation}
    S_o = f_o(S_p, \{h_i \mid D_i \in \text{Opponents}\}, H_r, T, g, w)
\end{equation}
where $f_p$ and $f_o$ represent the synthesis processes of proponents and opponents, respectively.

To ensure that the arguments from both the Proponents and Opponents align with the task and goals, we introduce a \textbf{judge}. At the end of each debate round, the judge decides which side has the strongest argument. If both sides meet the requirements, the debate ends; otherwise, it continues.
\begin{equation}
    (E_p, E_o) = J(S_p, S_o, H_r, T, g, P)
\end{equation}
where $E_p$ and $E_o$ are evaluations of strengths and weaknesses. The judge determines:
\begin{equation}
    W_r =
\begin{cases}
    S_p & \text{if } J \text{ deems } S_p \text{ optimal} \\
    S_o & \text{if } J \text{ deems } S_o \text{ optimal} \\
    \emptyset & \text{if no solution is optimal}
\end{cases}
\end{equation}
If $W_r \neq \emptyset$, output $W_r$ and terminate. Otherwise, proceed to round $r = r+1$.
If an optimal workflow is found:
\begin{equation}
    W = W_r
\end{equation}

If the maximum rounds $R$ are reached:
\begin{equation}
    W = J_{\text{final}}(\{S_p^{(r)}, S_o^{(r)} \mid r=1,2,\dots,R\}, H_R, T, g)
\end{equation}
where $J_{\text{final}}$ selects the best solution based on historical outcomes.

\textbf{Selection.} The framework starts with an empty initial workflow, which follows an input-output (IO). As we continuously evaluate each workflow in the environment, every workflow receives a score. To optimize each workflow, we use multi-agent debate. To avoid getting stuck in local optima, we employ a soft mixed probability selection strategy to choose which workflow to optimize at each step. The formula for this selection strategy is as follows:
\begin{equation}
P_{\text{mixed}}(i) = \lambda \cdot \frac{1}{n} + (1 - \lambda) \cdot \frac{\exp(\alpha \cdot (s_i - s_{\text{max}}))}{\sum_{j=1}^{n} \exp(\alpha \cdot (s_j - s_{\text{max}}))}
\end{equation}
Where $n$ is the number of candidate workflow, $s_i$ is the score of workflow $i$, $s_{\text{max}}$ is the maximum score, $\alpha$ controls the influence of scores, and $\lambda$ balances uniform and weighted probabilities.

\subsection{Fine-grained Feedback}
Previous approaches predominantly utilized coarse-grained feedback, which often resulted in suboptimal optimization across each layer. To address this, we have introduced fine-grained feedback to enhance the precision and effectiveness of the optimization process.

The optimized workflow is executed within the environment, logging any failures
for analysis. An error-correcting agent dissects the workflow to identify the failure-causing steps, generating an initial gene. This gene is refined using long/short
memory, which stores past gene records, to produce a dominant gene. The
dominant gene is then stored and logged in the failed workflow, providing a
reference for the next multi-agent debate.

To formalize this process, let $ g_0 $ be the initial gene generated by the error-correcting agent, and let $ M $ represent the long / short memory of the past genes. The dominant gene $ g $ can be expressed as:
\begin{equation}
    g = \mathcal{F}_{\text{LLM}}(g_0, M)
    \label{eq:gene-refinement}
\end{equation}
where $\mathcal{F}_{\text{LLM}}$ denotes the LLM-based gene refinement process that integrates $ g_0 $ with $ M $ to minimize failures.

\begin{table*}[ht]
\centering
\small	 
\begin{tabularx}{\textwidth}{l *{7}{>{\centering\arraybackslash}X}}
\toprule
\textbf{Method} & \textbf{MATH} & \textbf{HotpotQA} & \textbf{HumanEval} & \textbf{MBPP} & \textbf{ALFWorld} & \textbf{DROP} & \textbf{Avg.} \\
\midrule
GPT-4o-mini & 47.8 & 68.1 & 87.0 & 71.8 & 38.7 & 68.3 & 63.6 \\
CoT \cite{wei2022chain} & 48.8 & 67.9 & 88.6 & 71.8 & 39.9 & 78.5 & 65.9 \\
CoT SC \cite{wang2022self} & 47.9 & 68.9 & 88.6 & 73.6 & 40.5 & 78.8 & 66.4 \\
Self Refine \cite{Madaan2023SelfRefineIR} & 46.1 & 60.8 & 87.8 & 69.8 & 40.0 & 70.2 & 62.5 \\
\hline
LLM-Debate\cite{Du2023ImprovingFA} & 48.5 & 66.4 & 88.7 & 70.3 & 44.6 & 75.2 & 65.6\\
MultiPersona \cite{Wang2023UnleashingTE} & 50.8 & 69.2 & 88.3 & 73.1 & 39.1 & 74.4 & 68.8 \\
\hline
ADAS \cite{Hu2024AutomatedDO} & 43.1 & 64.5 & 82.4 & 53.4 & 47.7 & 76.6 & 61.3 \\
AFlow \cite{Zhang2024AFlowAA} & 53.8 & 73.5 & 90.9 & 81.4 & 59.2 & 80.3 & 73.2 \\
\hline
\textbf{DebFlow(Ours)} & \textbf{56.5} & \textbf{75.4} & \textbf{92.0} & \textbf{82.6} & \textbf{62.3} & \textbf{82.7} & \textbf{75.3} \\
\bottomrule
\end{tabularx}
\caption{Performance comparison of single-agent execution methods, Debate-optimized methods, and automated workflow optimization methods. Executed with GPT-4o-mini on divided test set, averaged over three trials.}
\label{table1}
\end{table*}

\section{Experiments}
\subsection{Experiment Setup}
\quad \textbf{Tasks and Benchmarks.} 
We conduct experiments on six representative tasks covering four domains:(1)reading comprehension, HotpotQA\cite{yang-etal-2018-hotpotqa}, DROP\cite{dua-etal-2019-drop}; (2)math reasoning, MATH\cite{Hendrycks2021MeasuringMP}; (3)code generation, HumanEval\cite{Chen2021EvaluatingLL} and MBPP\cite{Austin2021ProgramSW}; (4)embodied, ALFWorld\cite{Shridhar2020ALFWorldAT}.
Following prior studies such as \cite{Hu2024AutomatedDO}, we extracted 1,000 random samples each from the HotpotQA and DROP datasets. We also examined 617 problems from the MATH dataset, specifically choosing difficulty level 5 questions.

\begin{table}[ht]
\centering
\resizebox{\columnwidth}{!}{
\begin{tabular}{lccccc|c}
\toprule
\textbf{Method} & \textbf{MATH} & \textbf{HotpotQA} & \textbf{HumanEval} & \textbf{MBPP} & \textbf{DROP} & \textbf{Avg.} \\
\midrule
AFlow\textsubscript{gpt-4o-mini} & 4.76 & 5.12 & 0.84 & 5.56 & 3.36 & 3.93 \\
DebFlow\textsubscript{gpt-4o-mini} & \textbf{4.23} & \textbf{3.34} & \textbf{0.61} & \textbf{1.78} & \textbf{1.24} & \textbf{2.24} \\
\midrule
AFlow\textsubscript{deepseek} & 2.76 & 3.42 & 0.54 & 0.88 & 2.02 & 1.93 \\
DebFlow\textsubscript{deepseek} & \textbf{2.10} & \textbf{2.56} & \textbf{0.34} & \textbf{0.62} & \textbf{1.21} & \textbf{1.43} \\
\bottomrule
\end{tabular}
}
\caption{Training API costs. The baselines employ GPT-4o-mini as the optimizer, with subscripts indicating the model used as the executor.}
\label{table2}
\end{table}

\textbf{Baselines.} 
We compare DebFlow with three series of agentic baselines:(1) single-agent execution methods, IO (direct LLM invocation), Chain-of-Thought\cite{wei2022chain}, Self-Consistency (SC)\cite{wang2022self}, Self-Refine\cite{Madaan2023SelfRefineIR}; (2) Debate-optimized methods, MultiPersona\cite{Wang2023UnleashingTE}, LLM-Debate\cite{Du2023ImprovingFA}; (3) autonomous workflows, ADAS\cite{Hu2024AutomatedDO},  AFlow\cite{Zhang2024AFlowAA}.

\textbf{LLM Backbones.}
In our experimental framework, we utilize GPT-4o-mini-0718 as the optimizer, with all models accessed through APIs. The temperature is set to 1. We configured 20 iteration rounds for AFLOW, 10 for DebFlow, and 30 for ADAS.

\textbf{Evaluation Metrics.}
For quantitative assessment of model performance, we employ task-specific evaluation criteria across our experimental datasets. In mathematical reasoning tasks (GSM8K and MATHlv5 *), the accuracy of the solution is measured by the percentage metric of the solution rate. For programming proficiency evaluation (HumanEval and MBPP), we use the pass @1 metric, following the methodology established by \cite{Chen2021EvaluatingLL}. The performance of the question answering (HotpotQA and DROP) is assessed through the calculation of the F1 score.

\subsection{Experimental Results}
\quad \textbf{Main Results.}
Table \ref{table1} shows that DebFlow achieves superior performance by outperforming all three categories of existing methods, including single-agent execution methods, Debate-optimized methods, and automated workflow optimization approaches. Specifically, On the embodied benchmark ALFWorld, DebFlow achieves the optimal 62.3\%, outperforming the second-best AFLOW by 3.1\%.
On the MATH benchmark, it exceeds gpt-4o-mini by 8.7\% and surpasses the SOTA baseline AFlow by 2.7\%. Across six datasets in QA, Code, Embodied, and Math domains, Debflow surpasses all manually crafted workflows and demonstrates marginal improvements compared to automatically generated workflows.

\textbf{Cost Analysis.}
DebFlow demonstrates a highly resource-efficient agentic automation system. Using GPT-4o-mini as the optimizer and testing with the GPT-4o-mini executors and DeepSeek (three trials averaged), Table 2 highlights DebFlow's superiority, reducing costs by 43\% on average with GPT-4o-mini (e.g., 68\% savings on MBPP: 1.78$ vs. 5.56$ for AFlow) and 26\% with DeepSeek. This efficiency underscores DebFlow's ability to deliver superior results in varied LLM environments without excessive resource demands, making it practical for real-world deployments.



\begin{table}[h!t]
\centering
\small 
\renewcommand{\arraystretch}{1.2} 
\begin{tabularx}{\columnwidth}{l *{3}{>{\centering\arraybackslash}X}}
\toprule
\textbf{Task} & \textbf{w/o debate} & \textbf{w/o reflection} & \textbf{DebFlow} \\
\midrule
Math & 51.5 & 53.7 & 56.5 \\
\midrule
HotpotQA & 71.7 & 72.8 & 75.4 \\
\bottomrule
\end{tabularx}
\caption{The ablation study of DebFlow.}
\label{table3}
\end{table}


\textbf{Ablation Study.}
We perform an ablation study on two key components of DebFlow: (1) w/o Debate, removing the multi-agent debate mechanism and replacing it with a single LLM optimizer that generates candidate workflows randomly; and (2) w/o Reflection, removing the self-reflection module. We observe from Table \ref{table3} that removing the debate component causes the largest performance drop (-5.0\% on MATH and -3.7\% on HotpotQA), highlighting the importance of collaborative deliberation for reasoning. Removing the reflection module also degrades performance (-2.8\% on MATH and -2.6\% on HotpotQA), confirming its contribution to the overall effectiveness of DebFlow.

\textbf{Case Study.} As shown on the right of Figure \ref{total}, we present the workflows optimized by our framework on the MATH and HotpotQA benchmarks. HotpotQA tasks are often answerable directly from the questions, so optimal workflows are simple, requiring few operators. For the MATH dataset, we use level 5 difficulty questions. Simple workflows are often insufficient to solve these problems. Our framework can continuously optimize the workflows through multi-agent reasoning and fine-grained feedback, progressively adapting them to the tasks.

\section{CONCLUSION}
This study introduces DebFlow, a framework that uses multi-agent debate and fine-grained feedback to optimize workflows. Experiments show that DebFlow outperforms existing methods in performance and efficiency, while also reducing training resource consumption. In general, DebFlow provides an efficient, adaptable, and resource-friendly solution for automating agent creation.

\vfill\pagebreak

\small
\bibliographystyle{IEEEbib}
\bibliography{refs}

\end{document}